\documentclass[10pt,twocolumn,letterpaper]{article}

\usepackage[pagenumbers]{cvpr} 

\usepackage[dvipsnames]{xcolor}

\definecolor{cvprblue}{rgb}{0.21,0.49,0.74}
\usepackage[pagebackref,breaklinks,colorlinks,citecolor=cvprblue]{hyperref}

\usepackage{url}

\usepackage{algorithm}
\usepackage{algorithmic}
\usepackage{booktabs}
\usepackage{times}
\usepackage{epsfig}
\usepackage{amsmath}
\usepackage{amssymb}
\usepackage{subcaption}
\usepackage{multirow}
\usepackage[flushleft]{threeparttable}
\usepackage{pifont}

\usepackage{color}
\usepackage{array}
\usepackage{makecell}

\newlength\savewidth
\def\paperID{5235} 
\def\confName{CVPR}
\def\confYear{2024}

\newcommand{\YSR}[1]{\textcolor{blue}{#1}}

\title{A Large-Scale Analysis on Self-Supervised Video Representation Learning}

\author{Akash Kumar$^{1 \dagger} $  \qquad Ashlesha Kumar$^{2}$  \qquad Vibhav Vineet$^{3}$ \qquad  Yogesh Singh Rawat$^{1}$\\
CRCV, University of Central Florida$^{1}$ \qquad BITS Pilani$^{2}$ \qquad Microsoft Research$^{3}$\\
}

\begin{document}
\maketitle

\begin{abstract}
Self-supervised learning is an effective way for label-free model pre-training, especially in the video domain where labeling is expensive. Existing self-supervised works in the video domain use varying experimental setups to demonstrate their effectiveness and comparison across approaches becomes challenging with no standard benchmark. In this work, we first provide a benchmark that enables a comparison of existing approaches on the same ground. Next, we study five different aspects of self-supervised learning important for videos; 1) dataset size, 2) complexity, 3) data distribution, 4) data noise, and, 5) feature analysis. To facilitate this study, we focus on seven different methods along with seven different network architectures and perform an extensive set of experiments on 5 different datasets with an evaluation of two different downstream tasks. We present several interesting insights from this study which span across different properties of pretraining and target datasets, pretext-tasks, and model architectures among others. We further put some of these insights to the real test and propose an approach that requires a limited amount of training data and outperforms existing state-of-the-art approaches which use 10x pretraining data. We believe this work will pave the way for researchers to a better understanding of self-supervised pretext tasks in video representation learning.
\end{abstract}    
\section{Introduction}
\label{sec:intro}

\footnote{$^{\dagger}$Corresponding Author: Akash.Kumar@ucf.edu}

Deep learning models require a large amount of labeled data for their training. Obtaining annotations at large-scale needs a lot of effort and it becomes even more challenging as we shift from image to video domain. 
There are several interesting directions focusing on this issue such as
domain adaptation \cite{da}, knowledge distillation \cite{kd}, semi-supervised learning \cite{semi}, self-supervision \cite{main_survey} and weakly-supervised learning \cite{weakly}, which attempts to rely on
the knowledge learned from existing source datasets and transfer to new target datasets with minimal labels. 
Among these approaches, self-supervised learning use pretext task as supervisory signal and does not require any labels on source datasets 
which makes it more favorable.

 \begin{figure*}[t!]
     \centering
   \includegraphics[width=\linewidth]{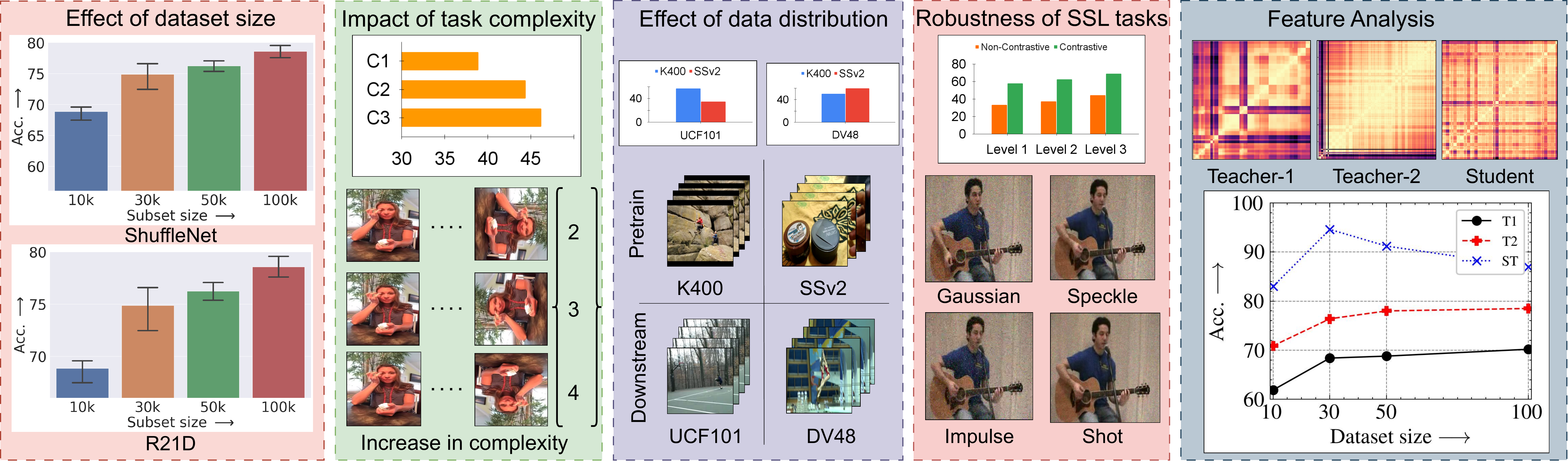}
     \caption{\textbf{Overview of proposed benchmark.} We study five different aspects in this benchmark study.
     Starting from left, 1) we show the analysis of \textit{effect of dataset size vs training time}. As the dataset size increases, variation in performance decreases even with longer training time, 2) We show the effect of task complexity (C1, C2, C3 - Different complexities). The bottom figure shows one use case of how complexity increases for the RotNet task, and, the top figure shows how the performance varies for the R21D network, 3) With different data distribution shifts, the third sub-figure shows the impact of \textit{target} data distribution on the \textit{source} data, 4) We look into another data distribution shift due to introduction of noise. We see how \textit{non-contrastive }tasks are more robust than \textit{contrastive} ones even with increasing levels of severity of noise. The bottom part shows an example for each type of noise. Clips are provided in supplementary, and, 5) Finally, we further analyze whether the features learn complimentary information. In this sub-figure, we show that using different architectures as teachers can substantially improve performance even in a low-data regime.
     }
\label{fig:main_fig}
\vspace{-15pt}
 \end{figure*}

In recent years, we have seen great progress in self-supervised learning (SSL) in video domain \cite{vcop, rotnet, prp, tdl, cvrl, rspnet}. 
More recently, the focus is more towards context-based learning which involves modifying input data such that to derive a classification \cite{pace, tclr, vcop, rotnet}, reconstruction \cite{prp, rspnet} or generative \cite{gen_1, gen_2, mem_dpc, videomae, videomoco} signal which can be used as a learning objective. 
The main focus of these works is designing a pretext task that is computationally inexpensive and which provides a strong supervisory signal such that the model learns meaningful \textit{spatio-temporal} features.

Despite this great progress, it is non-trivial to compare these approaches against each other due to a lack of standard protocols. 
These methods are evaluated under different conditions and there is no standard benchmark to evaluate the fair effectiveness of these methods.
A recent study \cite{Thoker2022HowSI} attempts to take a step towards this direction, but it is mainly focused on downstream learning, without exploring the self-supervision aspect which is one of the main goals in our study. 
In this work, we present a benchmark where important self-supervised pre-training parameters are kept consistent across methods for a fair comparison.
With the help of this benchmark, we study several critical aspects which are important for self-supervised learning; 
\textit{1) effect of pretraining dataset size, 2) task complexity, 3) generalization under distribution shift, 4) robustness against data noise, 5) properties of learned features. }

The proposed benchmark includes a large-scale assessment of context-based representative self-supervised methods for video representation learning. 
We analyze two different aspects: 1) \textit{learning objective} which includes \textit{contrastive} vs \textit{non-contrastive}, and 2) \textit{data transformation} that comprises three categories namely, \textit{spatial}, \textit{temporal}, and \textit{spatio-temporal}. We study seven different pretext tasks with seven different model architectures and
perform our experiments on five different video action recognition datasets and evaluate these approaches on two different downstream tasks, action recognition, and video retrieval.
We observe some interesting insights in this benchmark. Some of the key insights are; 1) Contrastive tasks are fast learners but are less robust against data noise,  2) there is no direct relation that increase in pretext task complexity leads to better understanding of spatio-temporal representation learning, 3) \textit{temporal} based pretext tasks are more difficult to solve than \textit{spatial} and \textit{spatio-temporal}, 4) spatio-temporal task can solve the pretext task independent of data distribution shifts, and finally, 5) we empirically show that these pretext tasks learn complementary features across factors such as model architecture, dataset distributions, dataset size, and pretext task.
Our contributions are threefold:
\begin{itemize}
    \item We present a benchmark for self-supervised video representation learning to compare different pretext tasks under a similar experimental setup.
    \item We perform extensive analysis on five important factors for self-supervised learning in videos; 1) dataset size, 2) task complexity, 3) distribution shift, 4) data noise, and, 5) feature analysis.
    \item Finally, we put some of our insights from this study to test and propose a simple approach that outperforms existing state-of-the-art methods on video action recognition with a limited amount of pretraining data. Additionally, based on our findings, we put down a set-up recipe for future self-supervised learning algorithms to build upon.
\end{itemize}

\section{Related Work}
\label{sec:related}

\paragraph{Self-supervised learning}
There are several works in the domain of self-supervised learning for video representation learning \cite{main_survey, schiappa2022self}. These approaches can be grouped into two main categories on the basis of pretext task:
1) context-based \cite{st_puzzles, context_2, vid_jigsaw, odd_one_out, pace, iic, visual_tempo, tclr, temporals, tdl, cvrl, rspnet, transrank, Guo2022CrossArchitectureSV, Ranasinghe2021SelfsupervisedVT}, and 2) cross-modal \cite{cross_modal_1, cross_modal_2, cross_modal_3}. 
Cross-modal approaches use multiple modalities such as audio, video, optical flow, and camera positions, and rely on consistencies across these modalities. Context-based learning exploits data transformations to derive supervisory signals for training the model. Context-based pretraining tasks have evolved a lot in the past few years. Our work explores the domain of how much variation in learned representations under different transformations. In contrast to other approaches, context-based approaches exploit the spatial and temporal information independently by several transformations \cite{shuffle_learn, odd_one_out, vcop, speednet, pace, cvrl, tdl}. Recent works have started to transform the spatial and temporal domain together \cite{st_puzzles, vcp, iic, prp, rspnet}. Incorporating multiple modalities improves performance, but, it's not available for all datasets, especially large-scale datasets. In this work, we restrict our focus to single-modality (RGB) approaches.
\vspace{-15pt}

\paragraph{Self-supervised benchmarking}
There are some prior efforts focusing on benchmarking self-supervised learning in the image domain. 
In \cite{goyal2019scaling}, the authors provide a detailed analysis of image-based self-supervised learning approaches and study how dataset size scaling affects the learned representations. Similarly in \cite{revisit}, the authors analyze how different model architectures play a role in visual self-supervised learning. In both these works, the authors did not focus on the importance of various pretext tasks themselves but only showed how certain pretext tasks can be improved. Therefore, their main focus was on downstream tasks rather than pretext learning. 
We, on the other hand, study different pretext tasks and analyze how various aspects affect feature learning. 
Moreover, these works are focused on the image domain, whereas we focus on the video domain. 
In recent work, \cite{unsupervised}, a study was performed to better understand unsupervised learning in the video domain, it basically explored the use of several pre-text tasks from the image domain and applied them to videos. 
We are not merely focusing on down-stream tasks and our attention is on the self-supervised aspect which includes factors 
such as data subset size, task complexity, dataset distribution, and noise robustness. 

\section{Self-Supervised Configurations}
We first describe the pretext tasks used in our study along with their categorization.
Then we discuss the details of this benchmark including
network architectures, datasets, downstream tasks and evaluations. 

\subsection{Tasks categorization}
We analyze two different aspects of video pretext tasks: 1) transformations applied to data, and 2) learning objectives. 
Data transformations include,
\textit{spatial-based (S)}, \textit{temporal-based (T)} and \textit{spatio-temporal (ST)}. 
\textit{Spatial} transformations include reshuffling of spatial patches, temporal consistent data augmentation, or rotation of images/patches. \textit{Temporal} tasks involve permutation classification of frames/clip, order verification, clips sampling at different paces, or, contrastive learning from temporal triplets. \textit{Spatio-temporal} tasks include those in which we modify both of these parameters simultaneously. 
This includes dilated sampling and simultaneous frame reconstruction, shuffling spatial and temporal domains, or, speed prediction, and contrastive visual features.
Learning objectives can be either \textit{contrastive} \cite{Chen2020ASF} or \textit{non-contrastive} such as \cite{videomae}.

Following this categorization, we select at least two representative pretext tasks from each \textit{transformation} category, one \textit{contrastive} and one \textit{non-contrastive}.
We study the following pretext tasks in this study; 
RotNet (Rot) \cite{rotnet}, Video Clip Order Prediction (VCOP) \cite{vcop}, Playback Rate Prediction (PRP) \cite{prp}, Spatiotemporal Contrastive Video Representation Learning (CVRL) \cite{cvrl}, Temporal Discriminative Learning (TDL) \cite{tdl}, Relative Speed Perception network (RSPNet) \cite{rspnet}, and \textit{V-MAE} \cite{videomae}. In concise summary,
1) \textit{RotNet} applies geometrical transformation on the data, 2) \textit{VCOP} learns the representation by predicting the permutation order, 3) \textit{PRP} has two branches, discriminative and generative that concentrate on temporal and spatial aspect respectively, 4) \textit{CVRL} learns to cluster the video of the same class with strong temporal coherent augmentations, 5) \textit{TDL} works on temporal triplets and minimizes the gap between anchor and positive on the basis of visual content, 6) \textit{RSPNet} applies contrastive loss in both spatial and temporal domain, and, 7) \textit{V-MAE} \cite{videomae} mask tokens of the input video and it tries to reconstruct those missing patches using an encoder-decoder architecture.
More details are provided in the supplementary. 

\subsection{Benchmark details}
\paragraph{Datasets:} 
We experiment with two different dataset types, 1) where appearance is more important, and 2) where time is more important. For appearance based, we use 
Kinetics-400 \cite{kinetics}, UCF101 \cite{ucf101}, and HMDB51 \cite{hmdb51}, where appearance is more important (recognize activity with a single frame) than temporal aspect, and for temporal aspect, we use Something Something-V2 \cite{ssv2} and Diving48 \cite{dv48}, where temporal information plays a significant role (require few frames to recognize activity). More details are in the supplementary.
\vspace{-5pt}

\paragraph{Spatio-temporal architectures} 
We analyze three different network capacities, 1) small-capacity, 2) medium-capacity, and 3) large-capacity. 
For small capacity, we study the following architectures; 
ShuffleNet V1 2.0X \cite{shufflev1}, SqueezeNet \cite{squeezenet}, and MobileNet \cite{mobnetv2}. For medium capacity we focus on conventional 3D architectures: C3D \cite{c3d}, R3D \cite{resnet3d}, and, R(2+1)D \cite{r21d} (R21D); . And, for big-capacity architectures, we study VideoSwin \cite{videoswin}, which is a transformer-based model. 
\vspace{-5pt}

\paragraph{Downstream tasks} We show results and analysis on two different downstream tasks - action recognition and clip retrieval. These two are the most prominent tasks in the field of self-supervised learning in videos. 
\vspace{-5pt}

\paragraph{Evaluation and analysis} 
We use top-1 accuracy for action recognition which indicates whether the class prediction is correct or not. Clip retrieval calculates the \textit{top-k} hits for nearest neighbor search, where $k=\{1, 5, 10, 20, 50\}$. For robustness performance, we calculate the relative robustness score $(R_{s})$ using original accuracy on clean test set $(A_{c})$ and perturbed accuracy on noisy test set$(A_{p})$ as $R_{s}= \frac{A_{c} - A_{p}}{A_{c}}$. We also provide qualitative feature analysis with the help of centered kernel alignment (CKA) maps \cite{cka}. CKA maps illustrate the model's hidden representations, finding characteristic block structures in models. There are two dominant properties of CKA maps: 1) \textit{Feature similarity: }  Lighter regions in map indicate more similar features between layers than darker regions. 2) \textit{Grid patterns: } Two main patterns stand out, a staggering grid, which indicates models are capable of learning more, and, distinctive light/dark block patterns meaning the network reached its saturation point.
\vspace{-5pt}

\section{Benchmark Analysis}

In this section, first, we perform some preliminary experiments to compare each pretext task under identical conditions. Then, we further perform analysis across the following five aspects in the next subsections. 

\noindent{\textit{\textbf{Effect of pretraining dataset size:}}}  
In self-supervised learning, a natural question to ask is whether dataset size plays any role in the performance of downstream tasks. It is important to study if the increase in the size of the pretraining dataset will proportionally reciprocate in performance improvement.
Also, a general trend is to train models for a very long duration at the pre-training stage. We investigate if the longer duration actually impacts the gain in performance. We look across different stages of training for multiple architectures and across different pretext tasks. 

\noindent{\textit{\textbf{Impact of task complexity:}}}
Some of the existing works show that increasing complexity leads to better representation learning, and if the complexity is decreased, the network will optimize to suboptimal solutions. 
We analyze this aspect in more detail with several tasks and different model architectures. 

\noindent{\textit{\textbf{Effect of data distribution:}}}
Existing self-supervised methods perform evaluations on K400 and UCF101 datasets. Both these datasets
fall into the same visual category with heavy appearance bias. However, we divert our attention towards datasets where the temporal dimension plays an important role such as SSv2 and Diving48. 

\noindent{\textit{\textbf{Robustness of SSL tasks:}}} In this aspect, we study the robustness qualities of SSL methods against data noise \cite{noise}.
We analyze which factors play a key role in the robustness of these methods against such distribution shifts. 

\noindent{\textit{\textbf{Feature analysis:}}} Finally, we look into feature space and analyze whether the learned representations are complimentary in nature when models are trained under different protocols.

 \begin{figure}[t!]
     \centering
   \includegraphics[width =\linewidth]{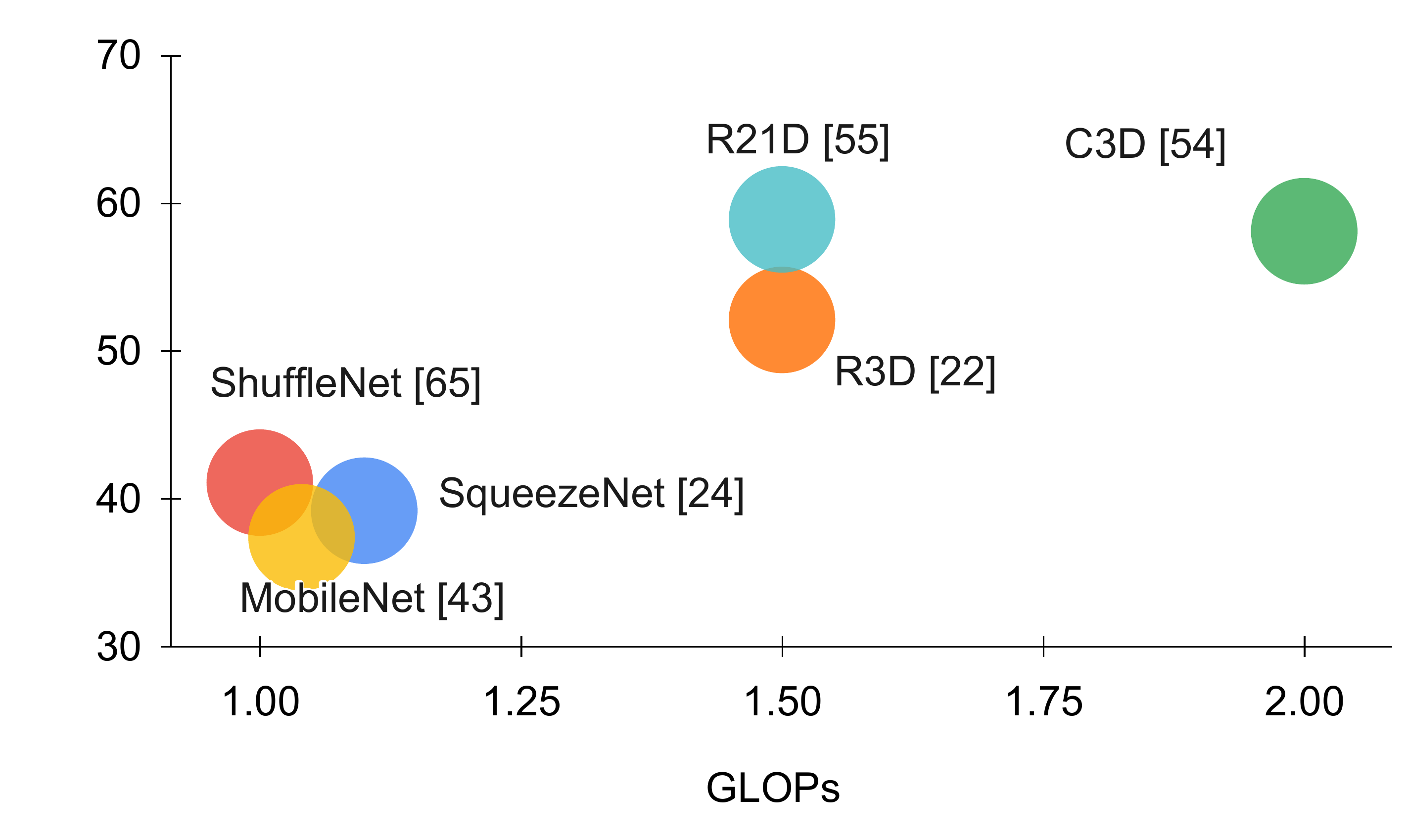}
     \caption{\textbf{Architecture Performance Analysis:} Variation in performance for different architectures. X-axis shows the relative floating point operations and Y-axis shows the Top-1 Accuracy.
     }
\label{fig:arch_var}
\vspace{-15pt}
 \end{figure}
 
\subsection{Preliminary Experiments}
\label{pre_exps}
First, we perform some preliminary experiments to analyze different architecture backbones, clip length, and evaluation with \textit{linear probing} vs \textit{finetuning}, and, finally layout discussion on the evaluation of different pretext tasks under the same constraints. 

\noindent \textbf{Backbone architectures: }Looking into smaller and medium capacity networks in Figure \ref{fig:arch_var}, ShuffleNet outperforms among smaller networks, whereas considering the trade-off between the number of trainable parameters and performance R21D performs better in medium network category.
Among big capacity networks, we look into a few recent end-to-end video-based transformer networks \cite{vivit, mvit, timesformer, videoswin}, and Video Swin \cite{videoswin} outperforms other architectures by a margin of 1-3\% on K400.

\noindent \textbf{Clip length: } Different pretext tasks take 16 or 32 frames as input clip length. We experimented with both 16 and 32 clips length and observe that 32 frames mostly provide better performance. However, to maintain consistency with most of the approaches and reduce computation costs, we use 16 frames in our experiments.

\noindent \textbf{Linear probe vs finetuning:} 
In the linear probe, we train only the linear layers attached for classification while freezing other network weights, whereas in finetuning the whole network is trained end-to-end. In our preliminary experiments we use Kinetics-400 for pretraining and UCF-101 as the target dataset. On several pretext tasks, we observe an average drop of 25\% (ShuffleNet) and 40\% (R21D) in performance when comparing linear probe with finetuning. 
However, we do not usually observe this significant drop when both the pretraining and target datasets are the same \cite{schiappa2022self}. It indicates that \textit{finetuning is important for the model to adapt to downstream dataset} in case it is different. 
Therefore, some of the existing works \cite{Thoker2022HowSI} rely on finetuning when the source and target datasets are different. 
Since we are interested in cross-dataset learning, we perform finetuning on all our downstream datasets.

\begin{table}[t!]
\footnotesize
\setlength{\tabcolsep}{4pt}
\centering
\begin{tabular}{c|cccc| ccc}
&\multicolumn{4}{c|}{Non-Contrastive} & \multicolumn{3}{c}{Contrastive}\\
 & Rot & VCOP & PRP & V-MAE & CVRL & TDL & RSP\\
 &(S) & (T) &(ST) & (ST) & (S) & (T) & (ST)\\
\hline
Shuffle  &16.6& 40.8  & 21.9 & -  & 62.3  & 12.4 & \textbf{68.8}  \\
R21D & 41.2  &  51.5 &46.2& 76.2 &61.2  &  31.7& \textbf{78.0}   \\
\textit{Reported $^{*}$} & 72.1 & 68.4 & 72.4 & 91.3 & \textbf{94.4 }& 84.9 & 93.7\\
\end{tabular}
\caption{\textbf{Comparison across different pretext tasks} pre-train on K400-50k subset and finetuned on UCF101 dataset against \textit{reported} results in the original paper.
}
\label{tab:compare_pretext}
\vspace{-25pt}
\end{table}

\noindent \textbf{Pretext tasks evaluation:}  
A comparison of pretext tasks on two different backbones is shown in Table \ref{tab:compare_pretext}.
We observe that
most of the \textit{contrastive} tasks outperform \textit{non-contrastive} tasks when they are trained under different constraints (row 3).
However, that is not the case when we compare them under the same constraints (row 1-2). 
Similarly, \textit{spatial} and \textit{spatio-temporal} tasks have a similar performance from reported results. However, \textit{spatio-temporal} pretext tasks outperform spatial ones by a large margin when we keep pre-training constraints similar. 
This supports our hypothesis that it is important to experiment under similar  constraints for a fair evaluation of different approaches. 
\subsection{Effect of dataset-size}
We first analyze the effects of pre-training data size variation. The network trains on four subsets of the K400 dataset: 10,000 (10k), 30,000 (30k), 50,000 (50k), and 100,000 (100k). The number of videos per class is the same. The smaller pre-training dataset is a subset of the bigger pre-training dataset size (i.e. $10k \subset 30k$ and so on). We look into three aspects regarding \textit{dependence on pre-train subset size:} a) behavior of different pretext tasks with the increase in pre-train dataset subset, b) performance across the different capacity of backbones, and, c) the effect of training time across different pretext tasks.

\begin{figure}[t!]

      \includegraphics[width=\linewidth]{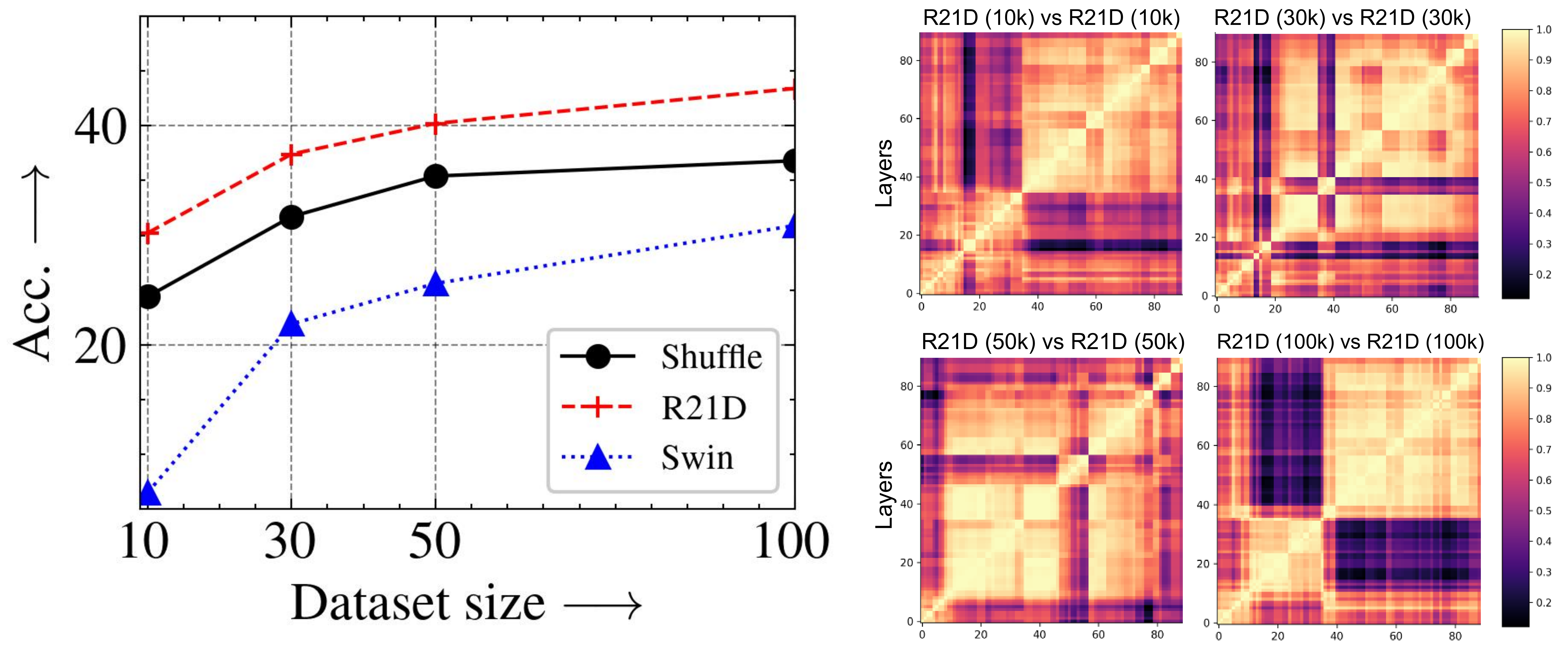}
  
    \caption{Left: \textbf{Dataset subset} performance for three different architectures on RSPNet pretext task (x-axis: subset size, y-axis: Top-1 Accuracy). Here, 10 means 10k dataset subset, 30 means 30k, and so on.  Right: \textbf{CKA maps} for RSPNet on different subsets with R21D backbone.
    }
    \label{fig:subset_analyze}
    \vspace{-10pt}
\end{figure}

\noindent \textbf{Observations: } From Table \ref{tab:pretrain_subset_tasks}, we observe that apart from TDL each pretext task performance improves with an increase in subset size. If we look into specific pretext task transformation category (Table \ref{tab:pretrain_subset_tasks}), the most gain with an increase in data is for \textit{spatio-temporal} tasks (~13\%), whereas the least gain is for \textit{temporal} pretext tasks (~3\%). Looking across different architectures in Figure \ref{fig:subset_analyze}, there's a minimal gain for R21D and ShuffleNet beyond increasing dataset size from 30k subset against VideoSwin which improves with an increase in dataset size which relates to similar behavior like image models discussed in \cite{goyal2019scaling}. Analyzing the effect of the duration of training across different pretext tasks, in Table \ref{tab:duration_all}, the performance gain is minimal ($<$1.5\%) after training for more than 100 epochs. Comparing contrastive and non-contrastive approaches, the gain in contrastive-based approaches is on average 1\% compared to 5\% for non-contrastive tasks beyond \textit{100 epochs} of training. 

\begin{table}[t!]
\small
        \centering
        \begin{tabular}{c|ccc |ccc}
        &\multicolumn{3}{c|}{Non-Contrastive} & \multicolumn{3}{c}{Contrastive}\\
        Subset & Rot &VCOP & PRP & CVRL & TDL & RSPNet  \\
        \hline
        10k &37.6 & 46.3  &17.5&55.9&31.1&70.9 \\
        30k & 36.2 & 50.4 &42.7 &56.9  & 30.9&76.4\\
        50k & 41.2 & 51.5 &46.2&61.2&30.2 &78.0\\
        \end{tabular}
        \caption{Evaluation of different pretext tasks on \textbf{different subset size} on R21D network.}
     \label{tab:pretrain_subset_tasks}
     \vspace{-8pt}
\end{table}


\begin{table}[t!]
\small
\centering
\begin{tabular}{c|ccc| ccc}
&\multicolumn{3}{c|}{Non-Contrastive} & \multicolumn{3}{c}{Contrastive}\\

Epochs & Rot & VCOP & PRP & CVRL & TDL & RSPNet  \\
\hline
50    & 35.4 &52.2&24.1&55.7 &32.1& 75.0\\
100  & 37.3&52.3&34.8& 58.5&31.3& 76.1\\
150  & 40.7&51.3&46.7& 60.2&31.5 &76.5\\
200  & 40.9& 52.8 & 45.0& 60.5&30.2& 77.4\\
\end{tabular}
\caption{\textbf{Performance at different stages} of training for all pretext tasks on R21D with 50k pre-training subset size.}
\label{tab:duration_all}
\vspace{-10pt}
\end{table}

\noindent \textbf{Inference: } (i) \textit{Spatio-temporal pretext tasks improve most with increment in dataset size and are most dependent on it than others since it involves transformation along both axes: appearance (spatial) and motion (temporal).}  (ii) \textit{Contrastive tasks are fast learners against non-contrastive and reach their potential in a relatively shorter duration.} 

\subsection{Impact of change in task complexity}
Next, we study the effect of task complexity. In this aspect, we analyze only non-contrastive tasks as it is non-trivial to define task complexity for contrastive-based approaches. 
We analyze three different complexities (C1, C2, C3) for each task.  
The variation in complexity for each task is briefly discussed as follows: a) \textit{RotNet}:  vary the number of rotations between 2 to 4, b) \textit{VCOP}: increase the number of shuffle clips from 3 to 5, and, c) \textit{PRP}: modify the dilation sampling rates from 2 to 4 classes. We investigate the following aspects here: a) does an increase in complexity means better spatio-temporal features learned at the pre-training stage? b) does the capacity of architecture plays any role?

\begin{table}[t!]
\small
\centering

\begin{tabular}{c |ccc}
TC$\downarrow$& S & T & ST \\
\hline
C1 & 20.1/48.3 & 41.6/\textbf{56.8} & \textbf{24.2}/38.9\\
C2 & \textbf{20.2}/\textbf{58.3} & \textbf{41.8}/54.8 & 18.1/44.4\\
C3 & 16.6/41.2 & 40.6/55.6 & 21.9/\textbf{46.2}\\
\end{tabular}
\caption{\textbf{Complexity Variation.} TC: Task complexity. Results are shown on UCF101 with ShuffleNet/R21D backbone.}
\label{tab:shuffle_complexity}
\vspace{-15pt}
\end{table}





\noindent \textbf{Observations:} From Table \ref{tab:shuffle_complexity}, comparing across rows we observe ShuffleNet performance doesn't improve much or degrade significantly if the complexity of the task is increased.  CKA maps show the structure transforms from staggering grids to a multi-block pattern indicating saturation with an increase in complexity. In between different categories of transformation, performance improves with complexity for the bigger model in the case of the \textit{spatio-temporal} task. Between ShuffleNet and R21D, R21D gives staggering grids against dark block patterns for ShuffleNet which shows the model can still learn better features. CKA maps are provided in the supplementary.

\noindent \textbf{Inference: } (i) \textit{Increase in pretext task complexity doesn't always reciprocate to better spatio-temporal feature learning. It is dependent on the pretext task and also the model capacity.} 
(ii) \textit{If higher complexity improves features learning, the model should also have the capacity, otherwise the task will be too difficult for the model to learn meaningful representations.}  

\subsection{Effect of dataset distribution}
Shifting our focus to datasets that have more hidden cues in the temporal aspect, we add pre-training on SSv2 and finetuning on Diving48 to our experiments. 
We answer the following questions in this section; a) does the categorization of pretext-task matter on \textit{source (pre-training)} and \textit{target (downstream)} datasets? b) what is the impact of \textit{source} dataset when the pretext task focuses only on a single task either \textit{spatial} or \textit{temporal}? 

\begin{figure}[t!]
    \centering
    \begin{subfigure}{0.45\linewidth}
       \centering
      \includegraphics[width=\linewidth]{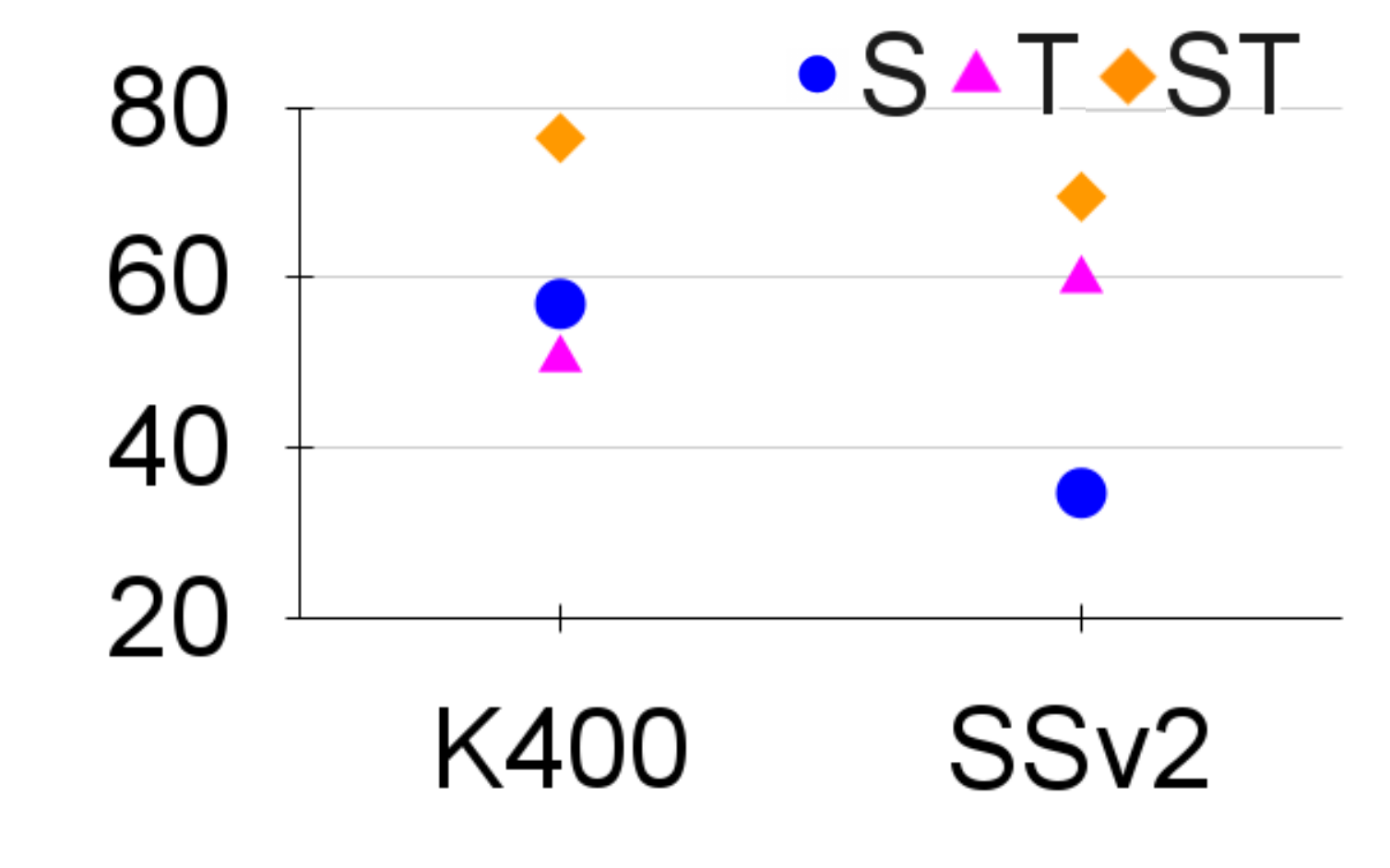}
      \caption{UCF101}
     \label{ood_ucf101}
    \end{subfigure}
    \begin{subfigure}{0.45\linewidth}
      \centering
     \includegraphics[width=\linewidth]{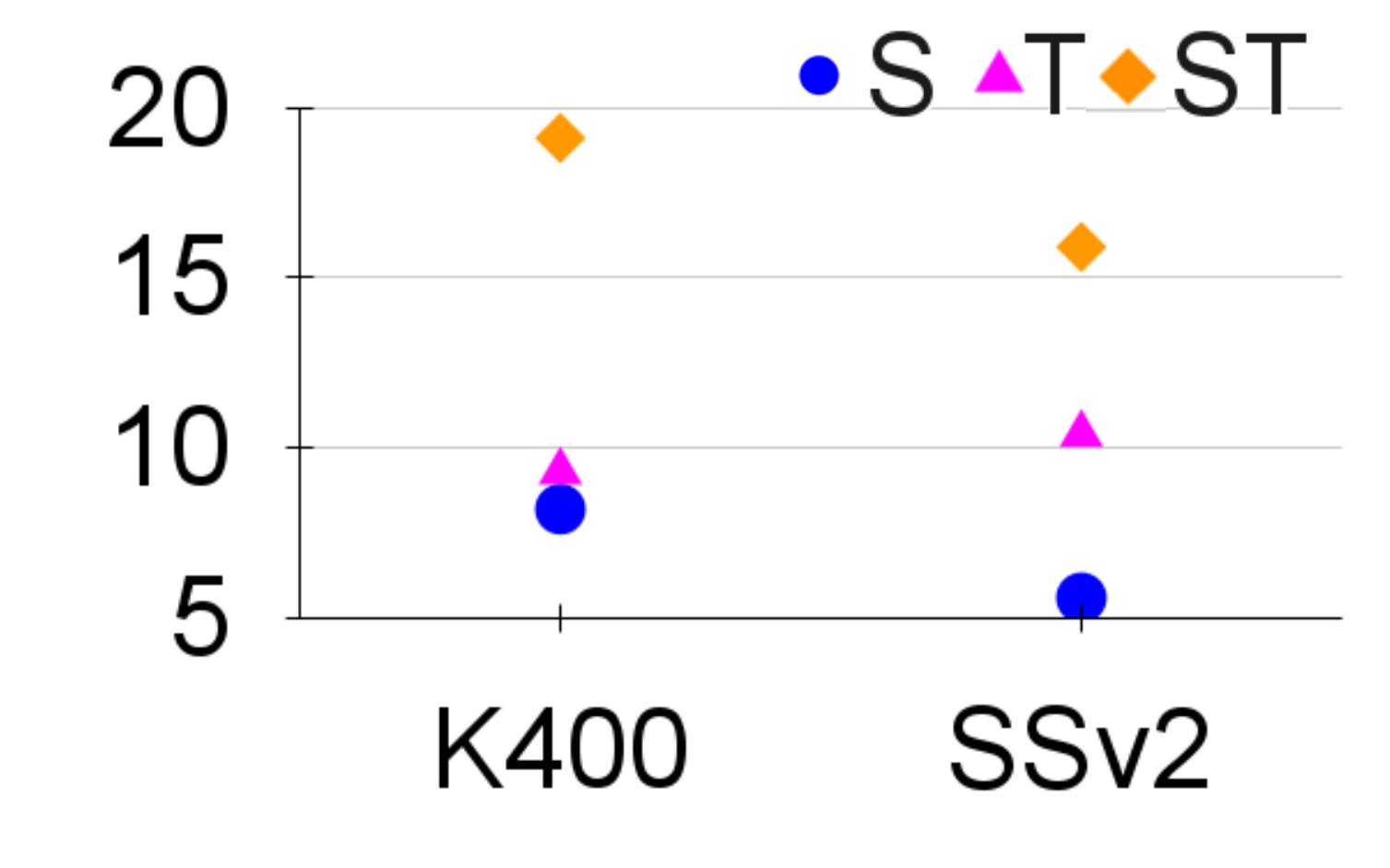}
     \caption{DV48}
     \label{ood_dv48}
    \end{subfigure}
    \caption{\textbf{Effect of different dataset distributions:} Pretraining on K400 and SSv2 with 30k subset size, finetuning on UCF101/Diving48 using R21D network. Here, S, T, and ST mean spatial(CVRL), temporal(VCOP), and, spatio-temporal(RSPNet) respectively. X-axis shows \textit{source} dataset and Y-axis shows Top-1 accuracy.
    }
    \label{fig:ood_multift}
    \vspace{-15pt}
\end{figure}

\noindent \textbf{Observations: } Looking into Figure \ref{fig:ood_multift}, we observe that \textit{spatio-temporal} pretext task outperforms other pretext tasks on both \textit{target} (downstream) datasets UCF101 and DV48 by a margin of 15-40\% and 10-13\% respectively whether the \textit{source} datasets is K400 or SSv2. Comparing, spatial and temporal-based pretext tasks, we see that they are \textit{majorly} dependent on \textit{source} datasets. Looking at Figure \ref{fig:ood_multift}, performance is better on both \textit{target} datasets if \textit{source} dataset has the same underlying properties as the pre-text task is trying to learn. Furthermore, the spatial task is more dependent on the \textit{source} dataset, since the relative drop on both UCF101 and DV48 for CVRL is significant (40\% and 30\% respectively) when the source dataset is SSv2 against K400. However, in the case of the temporal task, the drop is 15\% and 10\% respectively when the source dataset is K400 against SSv2.


\noindent \textbf{Inference: } (i) \textit{Spatio-temporal pretext task learns better features independent of source and target data distribution.} (ii) \textit{Spatial and temporal pre-text tasks are better learners when source data distribution belongs to spatial and temporal respectively.} (iii) \textit{Temporal pretext task prevails when target data is temporal, whereas, spatial is dependent on source data distribution.}

\subsection{Robustness of SSL tasks}

Similar to OOD datasets, introducing noise also shifts the distribution of datasets. We evaluate models on different types of noises introduced in \cite{ood_noise} with different severity levels on the UCF101 test dataset. Specifically, we probe into four different types of appearance-based noises: Gaussian, Shot, Impulse, and Speckle \cite{noise}. Here we look into the following aspects: a) how robust different categorizations of pretext tasks are? b) is the network's architecture dependent on the noise in the dataset? In the main paper, we only discuss one severity level and have provided a detailed analysis of multiple severity levels in the supplementary. 

\noindent \textbf{Observations: } From Table \ref{tab:r21d_perturb_l1}, we observe that the relative drop in performance for contrastive tasks is more than non-contrastive tasks for both R21D and ShuffleNet backbone. The most and least robust models are RotNet-R21D and PRP-R21D with 10.7\% and 70.1\% relative decrease. 


\begin{table}[t!]
\setlength{\tabcolsep}{4pt}
\small
        \centering
        \begin{tabular}{c| ccc| ccc |c}
        &\multicolumn{3}{c|}{Non-Contrastive} & \multicolumn{3}{c|}{Contrastive}\\
        & Rot & VCOP & PRP & CVRL & TDL & RSP & Avg. \\
        \hline
         R21D &10.7&  19.0 & 70.1 & 78.4&26.7 & 68.8 & 45.6\\
       Shuffle& 28.3 & 28.4 & 22.8 & 51.9 &43.5& 28.6 & 33.9\\
        \end{tabular}
        \caption{\textbf{Robustness analysis} on the relative decrease in \% performance across different pretext tasks on noisy UCF101 dataset. The performance is averaged over 4 noises.
        }
        \label{tab:r21d_perturb_l1}
        \vspace{-15pt}
\end{table}

\noindent \textbf{Inference:} \textit{Contrastive approaches are less robust to noise when compared with non-contrastive approaches.} 




\begin{figure*}[t!]
     \centering
     \small
   \includegraphics[width=\linewidth]{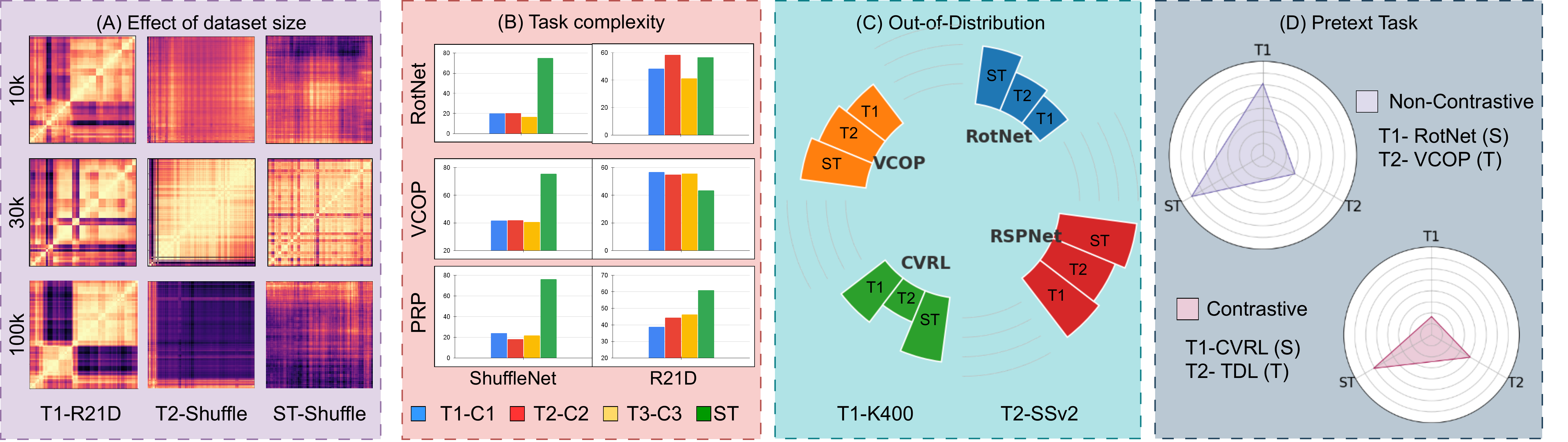}
     \caption{\textbf{Feature analysis overview.} This figure shows how knowledge distillation as a tool is beneficial across multiple scenarios. Brief details for each setup (Left to right): (A) \textit{Effect of dataset size:} Teachers (T1 and T2) are different architectures for a single subset. Student model (ST-Shuffle) CKA maps shows it learns complementary information especially for 30k. (B) \textit{Task Complexity: }  Teachers are multiple complexities across the same task. (C1, C2, C3 - different complexities as teachers.) We observe in most of the scenarios, Student  (ST) networks outperforms all teacher models which proves learning of orthogonal information from multiple teachers. (C) \textit{Out-of-Distribution: } Models from different \textit{source} datasets are teachers. Student model (ST) outperforms  both teachers trained on two different datasets. (D) \textit{Pretext Tasks: } Spatial and temporal task networks are teachers, and, student model (ST) learnt from two different categories of pretext tasks - spatial and temporal incorporate knowledge from both and outperforms both of the teachers for both contrastive and non-contrastive.}
\label{fig:kd_main_fig}
\vspace{-17pt}
 \end{figure*}
 
\subsection{Feature analysis}



We further analyze the learned features by these pretext tasks under different configurations. 
We specifically focus on understanding the complementary nature of these features. 
We employ knowledge distillation \cite{kdmain} as a tool to study this aspect. It is based on the idea that distilled knowledge from the ensemble of teacher networks makes the student model stronger. We use our benchmark models as teachers in different combinations to analyze whether a student learns orthogonal information on four different axes: 1) different architectures as the teacher within a \textit{dataset size}, 2) teachers with different complexities in a pretext task, 3) models from multiple \textit{source} datasets, and, 4) same architecture as teachers from multiple pretext tasks. Figure \ref{fig:kd_main_fig} summarizes the \textit{observations} for each aspect.

\begin{figure}[t!]
    \centering
    \begin{subfigure}{0.47\linewidth}
       \centering
      \includegraphics[width=\linewidth]{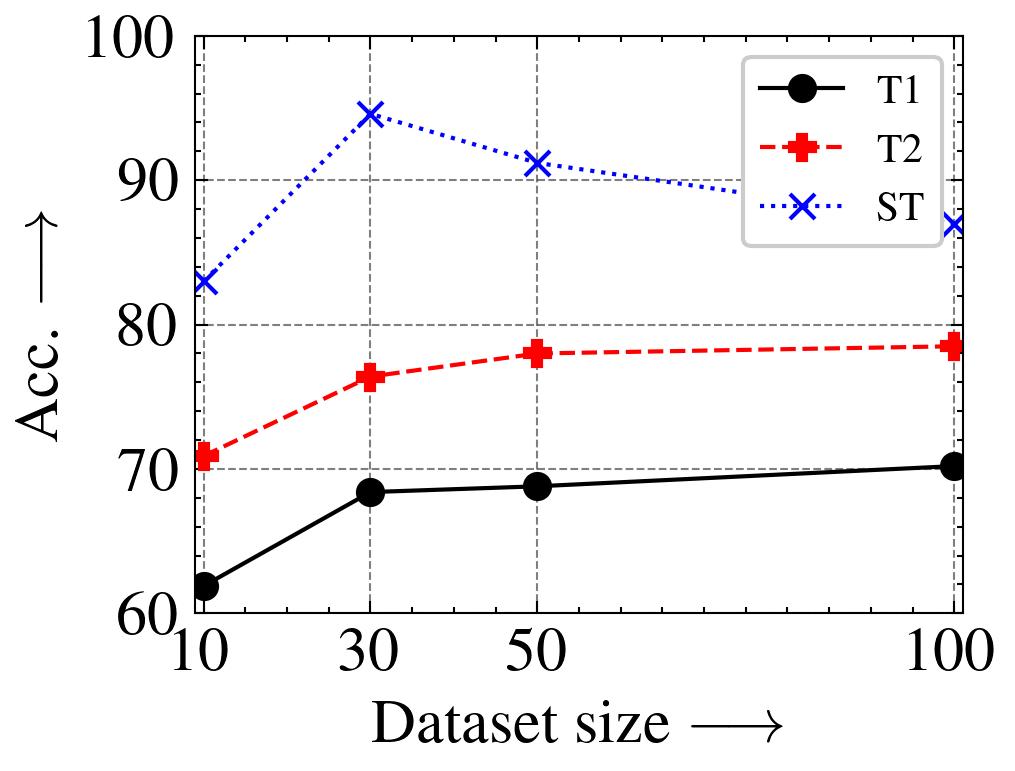}
    \caption{UCF101}
    \label{kd_ucf101}
    \end{subfigure}
    \begin{subfigure}{0.47\linewidth}
       \centering
      \includegraphics[width=\linewidth]{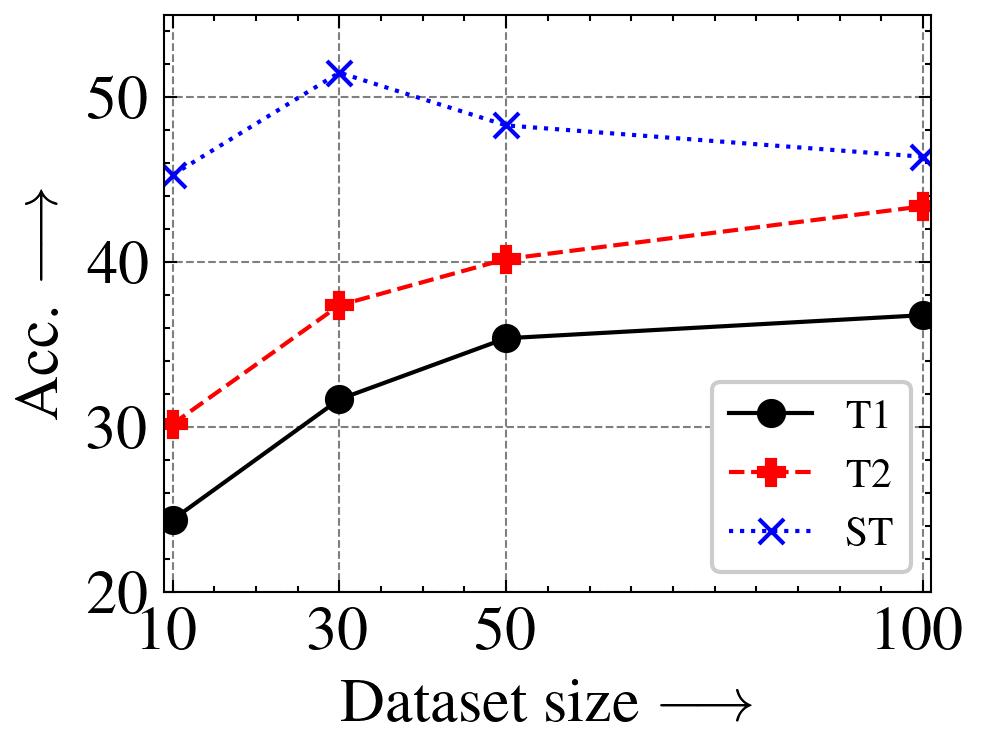}
    \caption{HMDB51}
    \label{kd_hmdb51}
    \end{subfigure}
    \caption{\textbf{Knowledge distillation} using teachers trained on multiple subset sizes on RSPNet. Student: ShuffleNet UCF101/HMDB51. Here T1 is Teacher-1 (shufflenet) and T2 is teacher-2 (R21D).
    }
    \label{fig:kd_dataset_size}
    \vspace{-20pt}
\end{figure}

\paragraph{Observations:} Although teacher network performance improves with subset, gain in complementary information reduces beyond 30k (Fig. \ref{fig:kd_dataset_size}). However, distillation does help in the reduction of training time with a significant improvement in performance which is evident from Fig. \ref{fig:kd_main_fig}(a). Independent of the pretext tasks category smaller architecture learns complimentary information and outperforms the teacher whereas bigger architecture it's task-dependent. Irrespective of task category whether transformation-based or contrastive, each task learns corresponding features from both source datasets and outperforms the teacher. Student network outperforms standalone spatio-temporal network performance in both contrastive and non-contrastive domains.

\noindent \textbf{Inference: }  (i) \textit{Knowledge can be distilled from different architectures for a given subset size (Fig. \ref{fig:kd_main_fig} (a))}, (ii) \textit{Knowledge from different \textit{source} datasets brings in complementary information (Fig. \ref{fig:kd_main_fig} (c))}, and (iii) \textit{Orthogonal features are learned across different categories of pretext tasks (Fig. \ref{fig:kd_main_fig} (d)).}

\section{Lessons Learned}

With all the analysis along studied axes, we learned a few lessons in-between these axes such as: (i) Contrastive tasks are fast learners but are also most susceptible to noise. (ii) An increase in dataset size or complexity does not help smaller models in learning better spatio-temporal features but these features are more robust to noise. (iii) Temporal tasks are relatively more difficult to learn since looking at the correlation between time of training, increase in dataset size, and complexity, the performance gain is minimal in each of this axis. It means this category of tasks is actually difficult to solve. (iv) Spatio-temporal pretext tasks improve with the increase in complexity and dataset size (if the model permits), and their behavior to learn better spatio-temporal features is independent of data distribution. 

Using these lessons, we further do more analysis in feature space. From there, we observe within an axis of comparison how models learn orthogonal information. Based on those observations, we analyze if we can push the performance for downstream tasks. We look into two downstream tasks: action classification and clip retrieval.

\begin{table}[t!]
\centering
\small
\begin{tabular}{l cccc}
\toprule
Approach& NxW/H & Backbone &  UCF101 & HMDB51 \\
\midrule
\textbf{Generative}\\
\midrule
VIMPAC \cite{Tan2021VIMPACVP}$^\dagger$ & 10x256 & ViT-L  & 92.7 & 65.9 \\
VideoMAE \cite{videomae}  & 16x224& ViT-B &91.3 & 62.6\\
VideoMAE $^{*}$ \cite{videomae} & 16x112& R21D-18 & 76.2 & 45.4\\
\midrule
\textbf{Context} \\
\midrule
PacePred \cite{pace} & 16x112& R21D-18 & 77.1 & 36.6\\
TempTrans \cite{temporals} &16x112 & R3D-18&  79.3& 49.8 \\
STS \cite{stats} & 16x112& R21D-18 & 77.8 & 40.5\\
VideoMoCo \cite{videomoco} &16x112& R21D-18 &  78.7 & 49.2\\
RSPNet \cite{rspnet} & 16x112& R21D-18 &  81.1 & 44.6\\
TaCo \cite{taco} &  16x224& R21D-18 & 81.8 & 46.0\\
TCLR\cite{tclr} &  16x112 & R21D-18 & 88.2 & 60.0\\
CVRL \cite{cvrl} &  32x224& R21D-18 &  92.9 &67.9\\
TransRank \cite{transrank}&  16x112 & R21D-18& 87.8 & 60.1 \\

\midrule
\textbf{Multi-Modal}\\
\midrule
AVTS \cite{avts}&   25x224 & I3D & 83.7 & 53.0 \\
GDT \cite{gdt}$^{\dagger}$ & 32x112& R21D &  95.2 & 72.8\\
XDC \cite{xdc} &  32x224& R21D &84.2 & 47.1\\
\midrule
Ours $^{*}$ & 16x112& R21D-18 & 97.3 & 51.5\\
\bottomrule
\end{tabular}
\caption{\textbf{Comparison with previous approaches} pre-trained on K400. Ours ( $^{*}$ best performing) is RSPNet pretrained on a 30k subset of K400. $^{\dagger}$ - Different pre-training data.}
\label{tab:action_recog_acc}
\vspace{-20pt}
\end{table}

\noindent{\textbf{Action Classification}} For this task, the model is finetuned end-to-end on downstream datasets, on UCF101 and HMDB51. In Table \ref{tab:action_recog_acc}, we compare our best-performing model  with other previous state-of-the-art approaches. \textit{\textbf{Observations:}} With only 30k videos compared to 200k+ videos used by other pretext tasks, we show that our model outperforms by a good margin on UCF101 against single and multi-modal approaches. We got competitive results on HMDB51 with a score of 51.5\%. 

\noindent{\textbf{Clip retrieval}} For this downstream task, we generate the feature vectors using pretraining weights. The nearest neighbor is found by measuring the cosine distance between test and train feature vectors. We show analysis on UCF101 and HMDB51, with different source data distributions, K400 and SSv2.  \textit{\textbf{Observations: }} Spatio-temporal task still outperform other categories independent of \textit{source} data distribution similar to what we observe earlier. Contrastive learns better \textit{appearance} features during the pre-training stage given both downstream datasets are \textit{appearance} based. Temporal tasks have almost similar performance pre-trained on either of the \textit{source} datasets, which shows even with an appearance-based dataset as a pre-train dataset, the task is not focusing much on spatial features. 
\begin{figure}[t!]
    \centering
    \begin{subfigure}{0.47\linewidth}
       \centering
      \includegraphics[width=\linewidth]{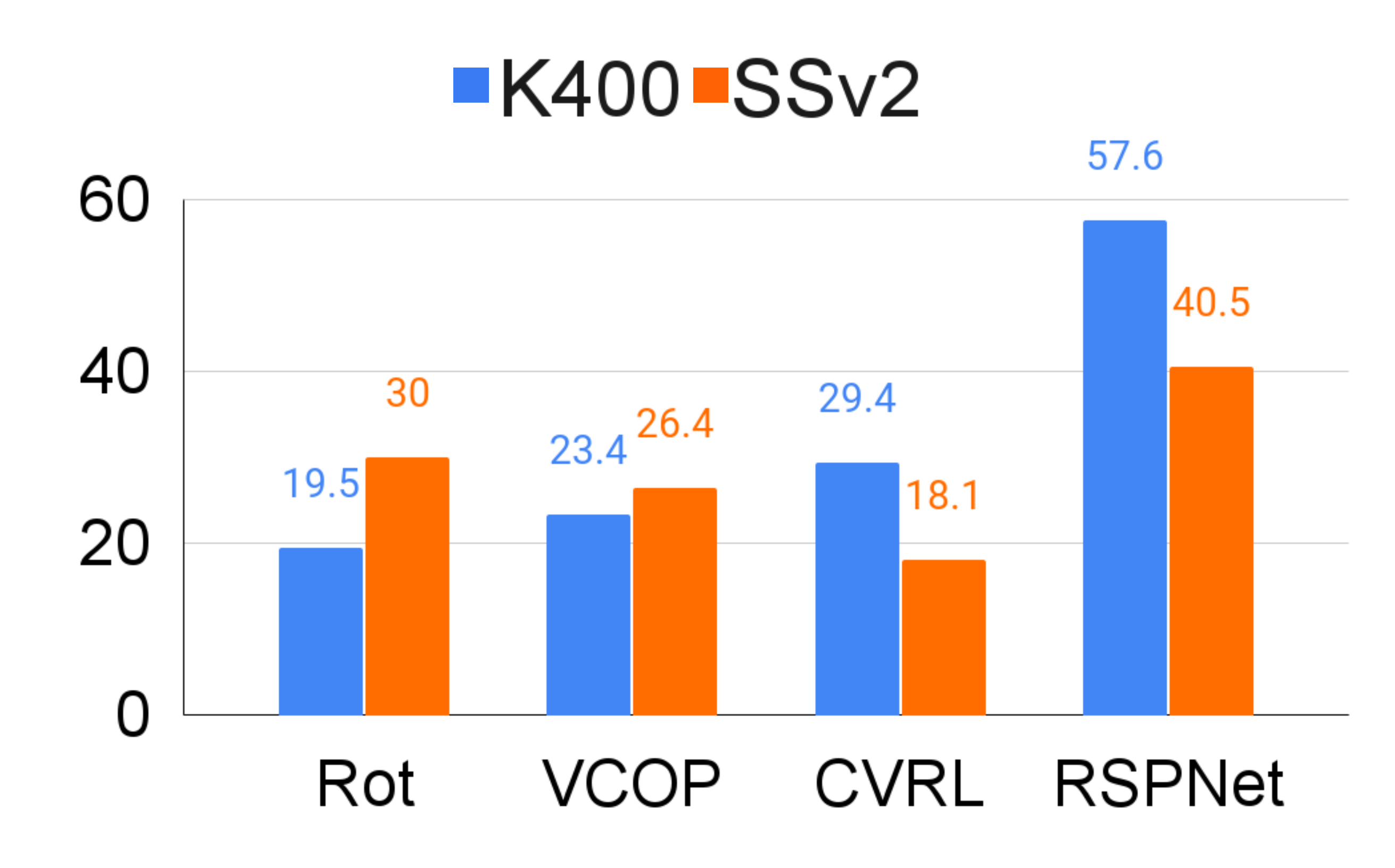}
    \caption{UCF101}
    \label{ret_ucf101}
    \end{subfigure}
    \begin{subfigure}{0.47\linewidth}
       \centering
      \includegraphics[width=\linewidth]{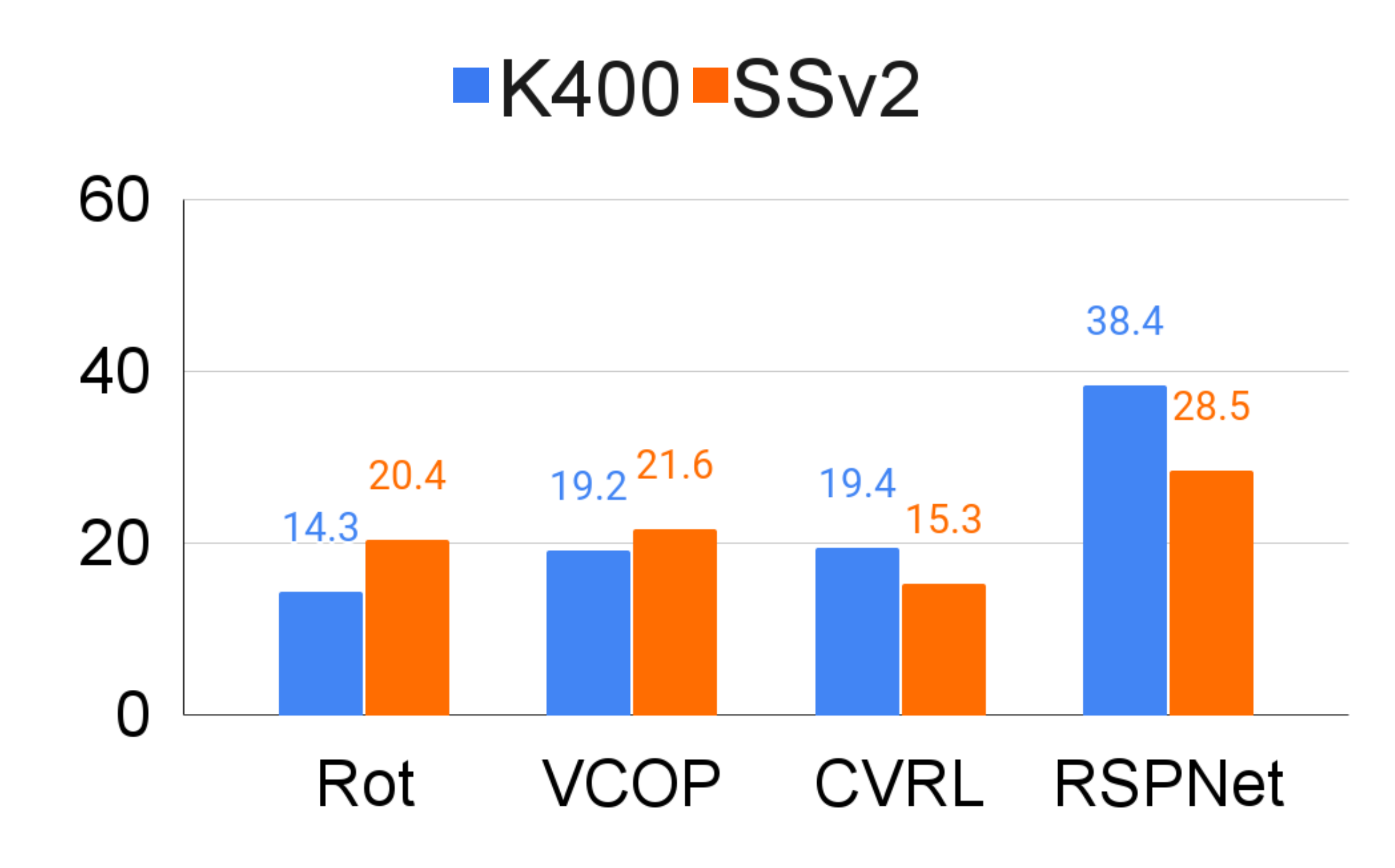}
    \caption{HMDB51}
    \label{ret_hmdb51}
    \end{subfigure}
    \caption{\textbf{Top@5 Clip Retrieval} - R21D on a) UCF101 and b) HMDB51, pre-trained on K400 and SSv2 - 30k subset.
    }
    \label{fig:clip_ret}
    \vspace{-20pt}
\end{figure}

\subsection{Surprising Findings}
We have multiple inference from different axes of analysis. However, to club a few which are new and helpful for video self-supervised community, we list down those here:

\noindent{\textit{\textbf{Dataset size and Training time Dependency:}}} Against the conventional belief that a lot of training data is a \textit{must} to achieve the best performance, we demonstrate that beyond a certain amount of training data, additional data provides diminishing returns for SSL in terms of performance improvement. This finding has significant implications, as it allows for a substantial reduction in the training data and there is almost a 10x reduction in training time which is particularly advantageous in computationally demanding video processing tasks. Furthermore, we show how KD as a tool, outperforms the original approach (100\% data) using almost 90\% less data further optimizing resource utilization by 80\%.
 
\noindent{\textit{\textbf{Robustness to real-world noise: }}}To our surprise, contrastive tasks are more susceptible to noise than non-contrastive ones. A smaller network tends to be more robust in some scenarios than a bigger network. We believe these findings are {\textit{novel and not known}} to the community as there is no existing study exploring these aspects and are helpful where robustness is necessary for real-world deployment.

\noindent{\textit{\textbf{Complementary knowledge:}}} Improvement in performance in the case of KD from different data distributions and categories of tasks brings out a recipe for a new SSL task. This involves utilizing a multi-teacher multi-student setup, where each teacher specializes in spatial and temporal tasks and is trained on a mixture of data sources. Our analysis indicates this would provide a powerful learning scenario.

\subsection{Recommendations}
Looking into several factors, here we provide some recommendations 
to set up the recipe for self-supervised learning:
1) \textit{Training speed:} If training time is a concern, contrastive tasks can help in reducing the pretraining time. The only downside is, they could be less robust against data noise.
2) \textit{Data distribution:} It is always better to use a spatio-temporal pretext task irrespective of the data distribution. However, if that is not an option, the pretext task should always be aligned with the nature of the pretraining dataset.
3) \textit{Model capacity:} If model capacity is limited, there is no benefit of increasing pretraining dataset size and using complex pretext tasks. 
4) \textit{Robustness:} If best performance is the goal we should use a non-contrastive as opposed to a contrastive pretext task.
5) \textit{Performance:} Pretext tasks learn complementary features across model architectures, pretraining datasets, pretext tasks, and tasks complexity, therefore, this complementary knowledge can be distilled to obtain strong spatio-temporal features.
\vspace{-10pt}

\section{Conclusion}
\label{sec:conclusion}
In this study, we explore different parameters for self-supervised learning in the video domain. We set a benchmark 
which provides an intuitive task categorization and 
enables a better comparison of different pretext tasks. 
Such an analysis has never been explored for video understanding to the best of our knowledge. 
We presented several interesting insights which will open up new directions for the research community. 
We also demonstrate the usefulness of some of these insights where we obtain state-of-the-art performance on video action recognition using merely a 10\% pretraining dataset when compared with existing methods. We believe this benchmark study will help the research community better understand self-supervised learning in the video domain. 
\vspace{-18pt} 
\section{Challenges and future work}

There are several key challenges in video SSL and we believe 1) long-term video understanding, 2) multi-modal learning, and 3) robust learning are some of the less studied aspects. 
The novel insights in our study regarding training dataset size, model architectures, and robustness will play a crucial role in guiding future work on these research directions. 
{
    \small
    \bibliographystyle{ieeenat_fullname}
    \bibliography{main}
}


\end{document}